\documentclass[10pt,twocolumn,letterpaper]{article}

\usepackage{cvpr}              % To produce the CAMERA-READY version
\usepackage[accsupp]{axessibility}
\usepackage[dvipsnames]{xcolor}

\definecolor{cvprblue}{rgb}{0.21,0.49,0.74}
\usepackage[pagebackref,breaklinks,colorlinks,citecolor=cvprblue]{hyperref}
\usepackage{multirow}
\usepackage{array}
\usepackage{siunitx}
\usepackage{tabularray}
\renewcommand{\paragraph}[1]{\vspace{2pt plus 1pt minus 1pt}\noindent{\bf #1}\;}
\def\paperID{3924} % *** Enter the Paper ID here
\def\confName{CVPR}
\def\confYear{2024}

\title{Text2Loc: 3D Point Cloud Localization from Natural Language}

\author{
	\small
	\begin{tabular}{c c c c c }                                            
		Yan Xia$^{*1,2 \footnotemark[2]}$ &
		Letian Shi$^{*1}$ &
		Zifeng Ding$^3$ &
		João F. Henriques$^4$ &
		Daniel Cremers$^{1,2}$ \\                                        
		\multicolumn{5}{c}{\shortstack{$^1$Technical University of Munich $^2$Munich Center for Machine Learning (MCML) \\ $^3$ LMU Munich $^4$ Visual Geometry Group, University of Oxford  }} \\                                                
		\multicolumn{5}{c}{\{yan.xia, letian.shi, cremers\}@tum.de, zifeng.ding@campus.lmu.de, joao@robots.ox.ac.uk } 
	\end{tabular}                                                                       
}

\begin{document}
\twocolumn[{%
\renewcommand\twocolumn[1][]{#1}%
\maketitle
\begin{center}
\vspace{-0.5cm}
\captionsetup{type=figure}
\includegraphics[width=0.99\linewidth]{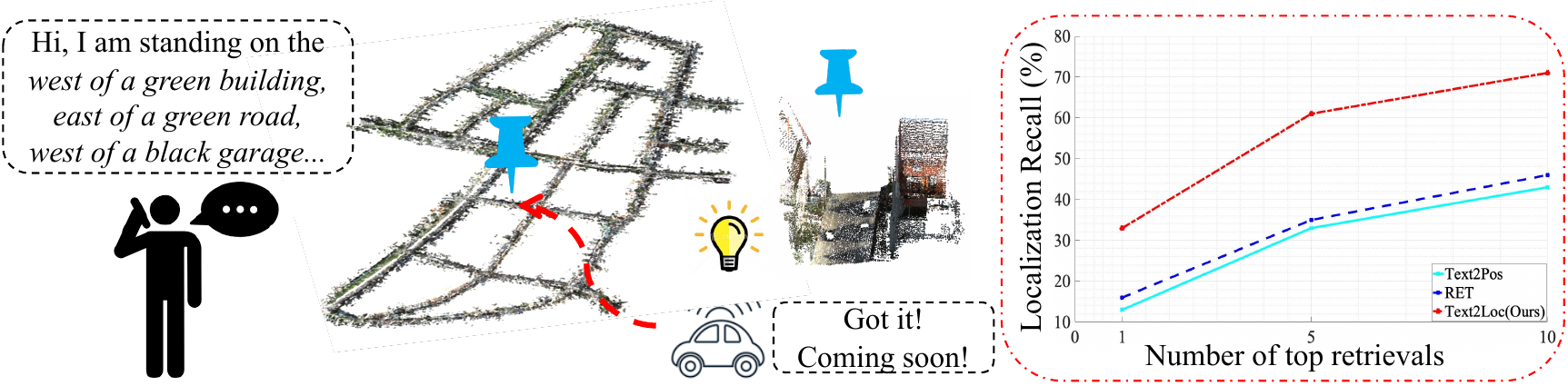}
% \vspace{-14pt}
\captionof{figure}{\textit{(Left)} We introduce Text2Loc, a solution designed for city-scale position localization using textual descriptions. When provided with a point cloud representing the surroundings and a textual query describing a position, Text2Loc determines the most probable location of that described position within the map. \textit{(Right)} Localization performance on the KITTI360Pose test set. The proposed Text2Loc achieves consistently better performance across all top retrieval numbers. Notably, it outperforms the best baseline by up to 2 times,  localizing text queries below \SI{5}{m}.}
\label{fig:teaser}
\end{center}%
}]

\let\oldthefootnote\thefootnote
\renewcommand{\thefootnote}{\fnsymbol{footnote}} 
\footnotetext[2]{Corresponding author. * Equal contribution.} 
\let\thefootnote\oldthefootnote

\begin{abstract}
We tackle the problem of 3D point cloud localization based on a few natural linguistic descriptions and introduce a novel neural network, Text2Loc, that fully interprets the semantic relationship between points and text. Text2Loc follows a coarse-to-fine localization pipeline: text-submap global place recognition, followed by fine localization.  In global place recognition, relational dynamics among
each textual hint are captured in a hierarchical transformer with max-pooling (HTM), whereas a balance between positive and negative pairs is maintained using text-submap contrastive learning. Moreover, we propose a novel matching-free fine localization method to further refine the location predictions, which completely removes the need for complicated text-instance matching and is lighter, faster, and more accurate than previous methods. Extensive experiments show that Text2Loc improves the localization accuracy by up to \textbf{$2\times$} over the state-of-the-art on the KITTI360Pose dataset. Our project page is publicly available at \url{https://yan-xia.github.io/projects/text2loc/}.

\end{abstract}    
\section{Introduction}
\label{sec:intro}
3D localization \cite{Min_2023_CVPR, Sarlin_2023_CVPR} using natural language descriptions in a city-scale map is crucial for enabling autonomous agents to cooperate with humans to plan their trajectories \cite{hu2023_uniad} in applications such as goods delivery or vehicle pickup~\cite{xia2023lightweight, xia2021vpc}.
When delivering a takeaway, couriers often encounter the ``last mile problem''. Pinpointing the exact delivery spot in residential neighborhoods or large office buildings is challenging since GPS signals are bound to fail among tall buildings and vegetation~\cite{Xia_2023_ICCV, xia2023perception}. Couriers often rely on voice instructions over the phone from the recipient to determine this spot. More generally, the ``last mile problem'' occurs whenever a user attempts to navigate to an unfamiliar place. It is therefore essential to develop the capability to perform localization from the natural language, as shown in Fig.~\ref{fig:teaser}.

As a possible remedy, we can match linguistic descriptions to a pre-built point cloud map using calibrated depth sensors like LiDAR. Point cloud localization, which focuses on the scene's geometry, offers several advantages over images. It remains consistent despite lighting, weather, and season changes, whereas the same geometric structure in images might appear vastly different.

The main challenge of 3D localization from natural language descriptions lies in accurately interpreting the language and semantically understanding large-scale point clouds. 
To date, only a few networks have been proposed for language-based localization in a 3D large-scale city map. Text2Pose~\cite{kolmet2022text2pos} is a pioneering work that aligns objects described in text with their respective instances in a point cloud, through a coarse-to-fine approach. In the coarse stage, Text2Pose first adopts a text-to-cell cross-model retrieval method to identify the possible regions that contain the target position. In particular, Text2Pose matches the text and the corresponding submaps by the global descriptors from 3D point clouds using PointNet++~\cite{qi2017pointnet++} and the global text descriptors using a bidirectional LSTM cell ~\cite{hochreiter1997long, 43895}. This method describes a submap with its contained instances of objects, which ignores the instance relationship for both points and sentences. Recently, the authors of RET~\cite{wang2023text} noted this shortcoming and designed Relation-Enhanced Transformer networks. While this results in better global descriptors, both approaches match global descriptors using the pairwise ranking loss without considering the imbalance in positive and negative samples.

Inspired by RET~\cite{wang2023text}, we also notice the importance of effectively leveraging relational dynamics among instances within submaps for geometric representation extraction. Furthermore, there is a natural hierarchy in the descriptions, composed of sentences, each with word tokens. We thus recognize the need to analyze relationships within (intra-text) and between (inter-text) descriptions. To address these challenges, we adopt a frozen pre-trained large language model T5~\cite{2020t5} and design a hierarchical transformer with max-pooling (HTM) that acts as an intra- and inter-text encoder, capturing the contextual details within and across sentences. Additionally, we enhance the instance encoder in Text2Pose~\cite{kolmet2022text2pos} by adding a number encoder and adopting contrastive learning to maintain a balance between positive and negative pairs. Another observation is that, when refining the location prediction in the fine localization stage, the widely used text-instance matching module in previous methods should be reduced since the noisy matching or inaccurate offset predictions are a fatal interference in predicting the exact position of the target. To address this issue, we propose a novel matching-free fine localization network. Specifically, we first design a prototype-based map cloning (PMC) module to increase the diversity of retrieved submaps. Then, we introduce a cascaded cross-attention transformer (CCAT) to enrich the text embedding by fusing the semantic information from point clouds. These operations enable one-stage training to directly predict the target position without any text-instance matcher.

To summarize, the main contributions of this work are:
\begin{itemize}

\item We focus on the relatively-understudied problem of point cloud localization from textual descriptions, to address the ``last mile problem''.

\item We propose a novel attention-based method that is hierarchical and represents contextual details within and across sentence descriptions of places.

\item We study the importance of positive-negative pairs balance in this setting, and show how contrastive learning is an effective tool that significantly improves performance.

\item We are the first to completely remove the usage of text-instance matcher in the final localization stage. We propose a lighter and faster localization model while still achieving state-of-the-art performance via our designed prototype-based map cloning (PMC) module in training and cascaded cross-attention transformer (CCAT).

\item We conduct extensive experiments on the KITTI360Pose benchmark~\cite{kolmet2022text2pos} and show that the proposed Text2Loc greatly improves over the state-of-the-art methods.

\end{itemize}

%-------------------------------------------------------------------------
\begin{figure*}[t!]
  \centering
\includegraphics[width=0.99\textwidth]
{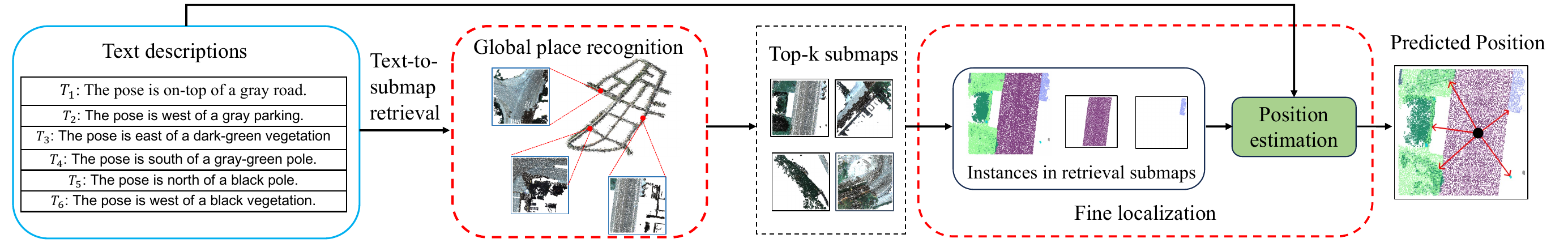}
\caption{The proposed Text2Loc architecture. It consists of two tandem modules: Global place recognition and Fine localization. \textit{Global place recognition.} Given a text-based position description, we first identify a set of coarse candidate locations, "submaps," potentially containing the target position. This is achieved by retrieving the top-k nearest submaps from a previously constructed database of submaps using our novel text-to-submap retrieval model. \textit{Fine localization.} We then refine the center coordinates of the retrieved submaps via our designed matching-free position estimation module, which adjusts the target location to increase accuracy.
}
\vspace{-1em}
  \label{fig:pipeline}
\end{figure*}

\section{Related work}
To date, only a few networks have been proposed for the natural language based 3D localization in a large-scale outdoor scene. Other tasks that are related to ours include 2D visual localization, 3D point cloud based localization, and 3D understanding with language.

\paragraph{2D visual localization.} Visual localization in 2D images has wide-ranging applications from robotics to augmented reality. Given a query image or image sequence, the aim is to predict an accurate pose. One of the early works, Scale-Invariant Feature Transform (SIFT)~\cite{lowe2004distinctive}, proposes the use of distinctive invariant features to match objects across different viewpoints, forming a basis for 2D localization. Oriented FAST and Rotated BRIEF (ORB)~\cite{rublee2011orb} has been pivotal in achieving robustness against scale, rotation, and illumination changes in 2D localization tasks. Recent learning-based methods~\cite{sarlin2019coarse, sattler2016efficient} commonly adopt a coarse-to-fine pipeline.  In the coarse stage, given a query image, place recognition is performed as nearest neighbor search in high-dimensional spaces. Subsequent to this, a pixel-wise correspondence is ascertained between the query and the retrieved image, facilitating precise pose prediction. However, the performance of image-based methods often degrades when facing drastic variations in illumination and appearance caused by weather and seasonal changes. Compared to feature matching in 2D visual localization, in this work, we aim to solve cross-model localization between text and 3D point clouds.

\paragraph{3D point cloud based localization.} With breakthroughs in learning-based image localization methods, deep learning of 3D localization has become the focus of intense research.  Similar to image-based methods, a two-step pipeline is commonly used in 3D localization: (1) place recognition, followed by (2) pose estimation. 
PointNetVlad~\cite{angelina2018pointnetvlad} is a pioneering network that tackles 3D place recognition with end-to-end learning. Subsequently, SOE-Net~\cite{xia2021soe} introduces the PointOE module, incorporating orientation encoding into PointNet to generate point-wise local descriptors. Furthermore, various methods~\cite{zhou2021ndt, deng2018ppfnet, fan2022svt, zhang2022rank, ma2022overlaptransformer, barros2022attdlnet, ma2023cvtnet} have explored the integration of different transformer networks, specifically stacked self-attention blocks, to learn long-range contextual features.
In contrast, Minkloc3D~\cite{komorowski2021minkloc3d} employs a voxel-based strategy to generate a compact global descriptor using a Feature Pyramid Network~\cite{lin2017feature} (FPN) with generalized-mean (GeM) pooling~\cite{radenovic2018fine}. However, the voxelization methods inevitably suffer from lost points due to the quantization step. CASSPR~\cite{Xia_2023_ICCV} thus introduces a dual-branch hierarchical cross attention transformer, combining both the advantages of voxel-based approaches with the point-based approaches. After getting the coarse location of the query scan, the pose estimation can be computed with the point cloud registration algorithms, like the iterative closest point (ICP)~\cite{segal2009generalized} or autoencoder-based registration~\cite{elbaz20173d}. By contrast to point cloud based localization, we use natural language queries to specify any target location.

\paragraph{3D vision and language.} Recent work has explored the cross-modal understanding of 3D vision and language. ~\cite{prabhudesai2019embodied} bridges language implicitly to 3D visual feature representations and predicts 3D bounding boxes for target objects. Methods~\cite{achlioptas2020referit3d, chen2020scanrefer, yuan2021instancerefer, feng2021free} locate the most relevant 3D target objects in a raw point cloud scene given by the query text descriptions. However, these methods focus on real-world indoor scene localization. Text2Pos~\cite{kolmet2022text2pos} is the first attempt to tackle the large city-scale outdoor scene localization task, which identifies a set of coarse locations and then refines the pose estimation. Following this, Wang \etal~\cite{wang2023text} propose a Transformer-based method to enhance representation discriminability for both point clouds and textual queries.

\section{Problem statement}
We begin by defining the large-scale 3D map ${ M_\textrm{ref} = {\left \{ m_{i}: i = 1,..., M \right \}}}$ to be a collection of cubic submaps $m_{i}$. Each submap ${ m_{i} = {\left \{ P_{i,j}: j = 1,..., p \right \}}}$ includes a set of 3D object instances $P_{i,j}$. Let \textit{${T}$}  be a query text description consisting of a set of hints $\{\Vec{h}_k\}_{k=1}^{h}$, each describing the spatial relation between the target location and an object instance. 
Following ~\cite{kolmet2022text2pos}, we approach this task in a coarse-to-fine manner. The text-submap global place recognition involves the retrieval of submaps based on $T$. This stage aims to train a function $F$, which encodes both $T$ and a submap $m$ into a unified embedding space. In this space, matched query-submap pairs are brought closer together, while unmatched pairs are repelled.
In fine-grained localization, we employ a matching-free network to directly regress the final position of the target based on $T$ and the retrieved submaps. 
Thus, the task of training a 3D localization network from natural language is defined as identifying the ground truth position $(x, y)$ (2D planar coordinates  w.r.t. the scene coordinate system) from \textit{${M_\textrm{ref}}$} :
\begin{equation} 
%    (x, y) = \phi\left(T,\; \underset{m\in M_\textrm{ref}}{{\rm argmin}}\: d\left(F(T), F(m)\right)\right),
\underset{\phi,F}{\min}\,\underset{(x,y,T)\sim\mathcal{D}}{\mathbb{E}}\left\Vert (x,y)-\phi\left(T,\;\underset{m\in M_{\textrm{ref}}}{{\rm argmin}}\:d\left(F(T),F(m)\right)\right)\right\Vert ^{2}
\label{Eq:encoder}
\end{equation}
\noindent 
where $d(\cdot, \cdot)$ is a distance metric (e.g. the Euclidean distance), $\mathcal{D}$ is the dataset, and $\phi$ is a neural network that is trained to output fine-grained coordinates from a text embedding $T$ and a submap $m$.

\section{Methodology}

Fig.~\ref{fig:pipeline} shows our Text2Loc architecture.
Given a text-based query position description, we aim to find a set of coarse candidate submaps that potentially contain the target position by using a frozen pre-trained T5 language model~\cite{2020t5} and an intra- and inter-text encoder with contrastive learning, described in Section~\ref{section: coarse stage}.
Next, we refine the location based on the retrieved submaps via a designed fine localization module, which will be explained in Section~\ref{section: fine stage}.
Section~\ref{section:loss function} describes the training with the loss function.

\subsection{Global place recognition}
\label{section: coarse stage}

\begin{figure}[t!]
  \centering
\includegraphics[width=1.0\linewidth]{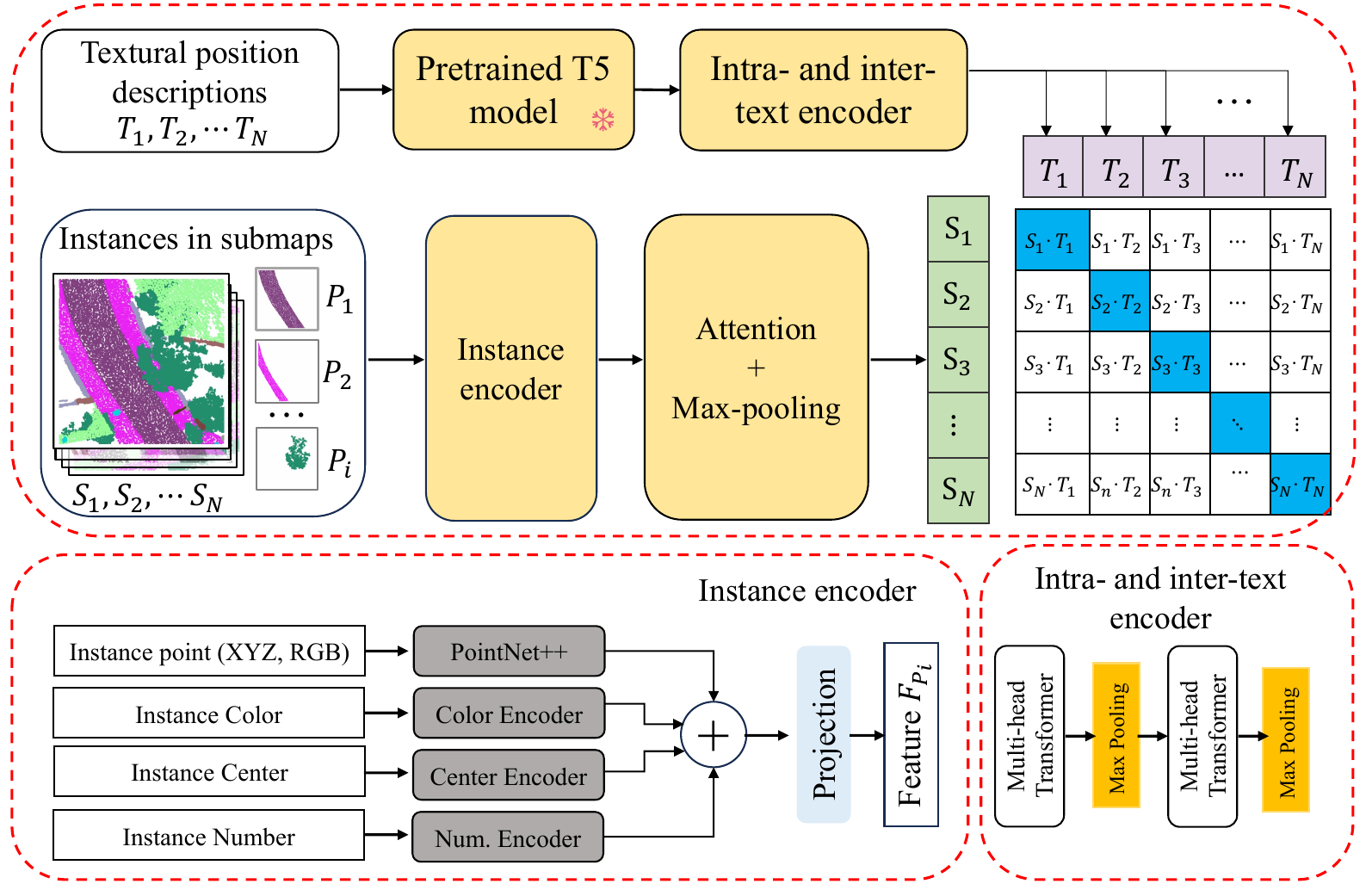}
  \caption{\textit{(top)} The architecture of global place recognition, \textit{(bottom)} instance encoder architecture for point clouds, and the architecture of intra- and inter-text encoder. Note that the pre-trained T5 model is frozen.}
  \label{fig:coarse}
  \vspace{-2em}
\end{figure}

3D point cloud-based place recognition is usually expressed as a 3D retrieval task. Given a query LiDAR scan, the aim is to retrieve the closest match and its corresponding location from the database by matching its global descriptor against the global descriptors extracted from a database of reference scans based on their descriptor distances.
Following this general approach, we adopt the \textit{text-submap} cross-modal global place recognition for coarse localization. With this stage, we aim to retrieve the nearest submap in response to a textual query. The main challenge lies in how to find simultaneously robust and distinctive global descriptors for 3D submaps $S$ and textual queries $T$. Similar to~\cite{kolmet2022text2pos,wang2023text}, we employ a dual branch to encode $S$ and $T$ into a shared embedding space, as shown in Fig.~\ref{fig:coarse} (top).

{\bf Text branch.} We initially use a frozen pre-trained large language model, T5~\cite{2020t5}, to extract nuanced features from textual descriptions, enhancing the embedding quality. We then design a hierarchical transformer with max-pooling layers to capture the contextual details within sentences (via self-attention) and across them (via the semantics that are shared by all sentences), as depicted in Fig.~\ref{fig:coarse} (Bottom right). Each transformer is a residual module comprising Multi-Head Self-Attention (MHSA) and FeedForward Network (FFN) sublayers. The feed-forward network comprises two linear layers with the ReLU activation function. More details are in the Supplementary Materials.

{\bf 3D submap branch.} Each instance $P_i$ in the submap $S_N$ is represented as a point cloud, containing both spatial and color (RGB) coordinates, resulting in 6D features (Fig.~\ref{fig:coarse} (bottom left)).  
We utilize PointNet++~\cite{qi2017pointnet++} (which can be replaced with a more powerful encoder) to extract semantic features from the points. 
Additionally, we obtain a color embedding by encoding RGB coordinates with our color encoder and a positional embedding by encoding the instance center $\Bar{P_i}$ (i.e., the mean coordinates) with our positional encoder. 
% Considering that different instances may contain varying numbers of points, we also design a number encoder to handle this variation by encoding the number of points in this instance.
We find that object categories consistently differ in point counts; for example, roads typically ($>1000$ points)  have a higher point count than poles ($<500$ points). We thus design a number encoder, providing potential class-specific prior information by explicitly encoding the point numbers.
All the color, positional, and number encoders are 3-layer multi-layer perceptrons (MLPs) with output dimensions matching the semantic point embedding dimension. We merge the semantic, color, positional, and quantity embeddings through concatenation and process them with a projection layer, another 3-layer MLP. This projection layer produces the final instance embedding ${F}_{p_i}$. 
Finally, we aggregate \textit{in-submap} instance descriptors $\{{F}_{p_i}\}^{N_p}_{i=1}$ into a global submap descriptor ${F}^S$ using an attention layer~\cite{xia2021soe} followed by a max pooling operation.

{\bf Text-submap Contrastive learning.} We introduce a cross-modal contrastive learning objective to address the limitations of the widely used pairwise ranking loss in \cite{kolmet2022text2pos, wang2023text}. This objective aims to jointly drive closer the feature centroids of 3D submaps and the corresponding text prompt. In our overall architecture, illustrated in Figure \ref{fig:coarse}, we incorporate both a text encoder and a point cloud encoder. These encoders serve the purpose of embedding the text-submap pairs into text features denoted as $F^T\in\mathcal{R}^{1\times C}$ and 3D submap features represented as $F^S\in\mathcal{R}^{1\times C}$, respectively. Here, $C$ signifies the embedding dimension.
Inspired by CLIP~\cite{radford2021learning}, we computer the feature distance between language descriptions and 3D submaps with a contrastive learning loss (See Sec.~\ref{section:loss function} for details).

\subsection{Fine localization}
\label{section: fine stage}

\begin{figure}[t!]
  \centering
  \includegraphics[width=1.0\linewidth]{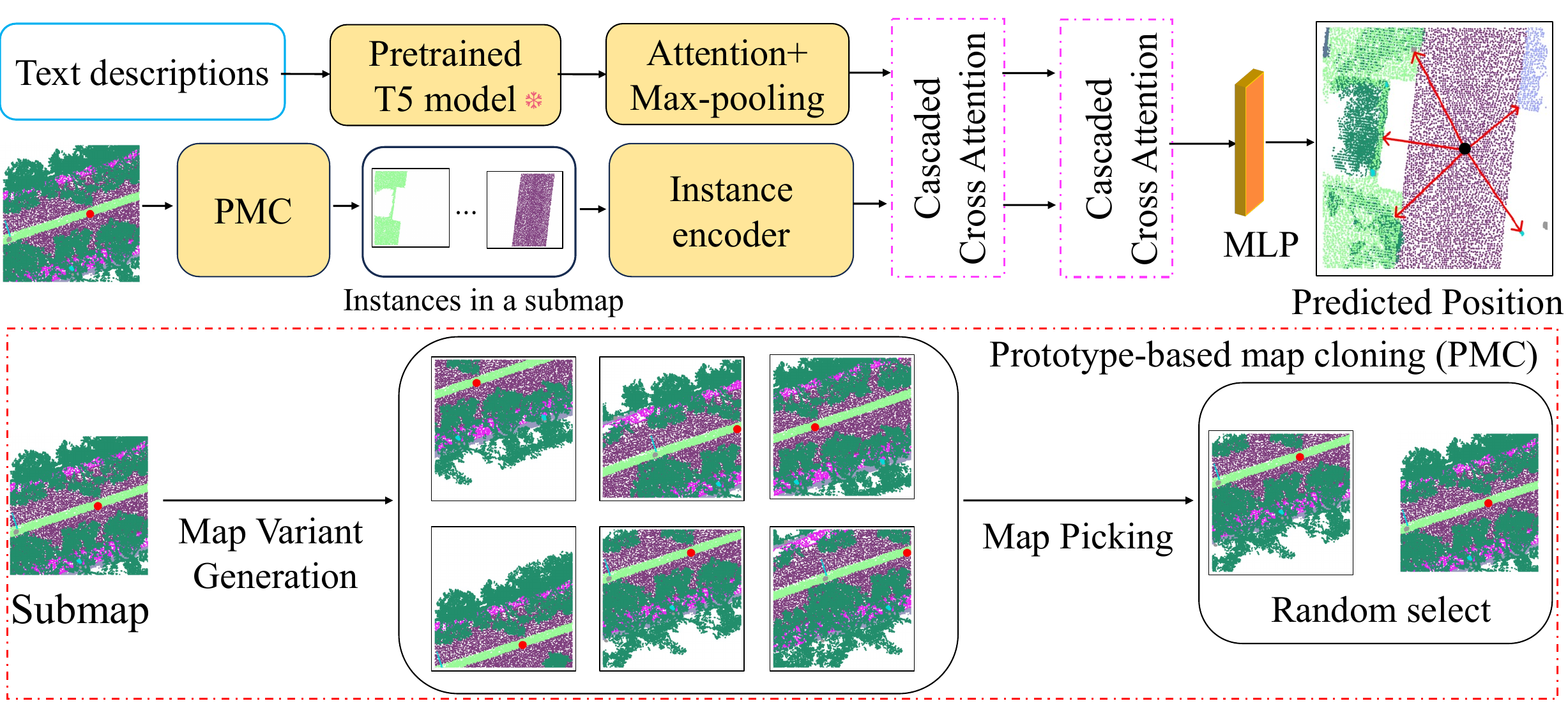}
    % \vspace{-7pt}
  \caption{The proposed matching-free fine localization architecture. It consists of two parallel branches: one is extracting features from query text descriptions \textit{(top)} and another is using the instance encoder to extract point cloud features \textit{(bottom)}. Cascaded cross-attention transformers (CCAT) use queries from one branch to look up information in the other branch, aiming to fuse the semantic information from point clouds into the text embedding. The result is then processed with a simple MLP to directly estimate the target position.}
  \label{fig:fine-loc}
  \vspace{-1em}
\end{figure}

Following the text-submap global place recognition, we aim to refine the target location prediction within the retrieved submaps in fine localization. 
% Their initial process involves aligning hints with instances, forecasting the offsets from each aligned instance's center point to the target, and ultimately averaging the predicted positions of all hints. 
Although the final localization network in previous methods~\cite{kolmet2022text2pos, wang2023text} achieved notable success using a text-submap matching strategy, the inherent ambiguity in the text descriptions significantly impeded accurate offset predictions for individual object instances. To address this issue, we propose a novel matching-free fine localization network, as shown in Fig.~\ref{fig:fine-loc}. The text branch (top) captures the fine-grained features by using a frozen pre-trained language model T5~\cite{2020t5} and an attention unit followed by a max pooling layer. The submap branch (bottom) performs a prototype-based map cloning module to increase more map variants and then extracts the point cloud features using an instance encoder, the same as in the global place recognition. We then fuse the text-submap feature with a Cascaded Cross-Attention Transformer and finally regress the target position via a simple MLP.

 % As now we estimate the absolute position in the submap, we need a better targets distribution in our training process among the submaps, so we apply Prototype-based Map Cloning to rectify our target's distribution.

{\bf Cascaded Cross-attention Transformer (CCAT).} To efficiently exploit the relationship between the text branch and the 3D submap branch, we propose a CCAT to fuse the features from the two branches. The CCAT consists of two Cross Attention Transformers (CAT), each is the same as in ~\cite{Xia_2023_ICCV}. The CAT1 takes the point cloud features as Query and the text features as Key and Value. It extracts text features with reference to the point features and
outputs point feature maps that are informed by the text features. Conversely, CAT2 produces enhanced text features by taking the text features as the Query and the enhanced point cloud features from CAT1 as the Key and Value. Notably, the CAT1 and the CAT2 are a cascading structure, which is the main difference from the HCAT in ~\cite{Xia_2023_ICCV}. In this work, two cascaded CCATs are used. More ablation studies and analyses are in the Supplementary Materials. 

{\bf Prototype-based Map Cloning (PMC).} To produce more effective submap variants for training, we propose a novel prototype-based map cloning module. For each pair $\{T_i, S_i\}$, we hope to generate a collection $ \mathcal{G}_i$ of surrounding map variants centered on the current map $S_i$, which can be formulated as follows: 

\begin{equation}
\begin{aligned}
    \mathcal{G}_i = \{S_j \; | \; & \big \| {\Bar{s_j} - \Bar{s_i}} \big \|_{\infty}< \alpha ,
    \; \big \|\Bar{s_j} - c_i \big \|_{\infty} < \beta \; \},
    % \\
    % & \text{Missing}(S_j, T_i) \leq N_{m}, \; i \neq j \}
\end{aligned}
\end{equation}
\noindent
where $\Bar{s_i}$, $\Bar{s_j}$ are the center coordinates of the submaps $S_i$ and $S_j$ respectively.
$c_i$ represents the ground-truth target position described by $T_i$, $\alpha$ and $\beta$ are the pre-defined thresholds. In this work, we set $\alpha=15$ and $\beta=12$.

In practice, we find that certain submaps in $ \mathcal{G}_i$ have an insufficient number of object instances corresponding to the textual descriptions $T_i$. To address this, we introduce a filtering process by setting a minimum threshold $N_{m}=1$. This threshold implies that at most one instance mismatch is permissible. After applying this filter, we randomly selected a single submap from the refined $ \mathcal{G}_i$ for training.
% The neighbor set consists of submaps adjacent to the original one, encompassing the described target from the text, and falling below the threshold for missing items. 
%Missing$(S, T)$ defines the number that submap S misses the instances mentioned in the hints of the text T. 

% $\text{Missing}(S_j, T_i) \leq N_{m}$ indicates that, if a submap $S$ lacks more than $N_m$ instances corresponding to textual descriptions, then we remove this submap from the neighbor set.

\subsection{Loss function}
\label{section:loss function}

{ \bf Global place recognition.} Different from the pairwise ranking loss widely used in previous methods~\cite{kolmet2022text2pos, wang2023text}, we train the proposed method for text-submap retrieval with a cross-model contrastive learning objective. Given an input batch of 3D submap descriptors  $\{{F}^S_i\}^{N}_{i=1}$ and matching text descriptors $\{{F}^T_i\}^{N}_{i=1}$ where $N$ is the batch size, the contrastive loss among each pair is computed as follows,
% \begin{equation}\footnotesize
%       l(i,T,S) = -\log\frac{\exp(F^T_i\cdot F^S_{i}/\tau)}{\exp(F_i^T\cdot F^S_i/\tau) + \sum\limits_{j\in N, j\neq i} \exp(F_i^T\cdot F_j^S/\tau)},
% \end{equation}
\begin{equation}\footnotesize
      l(i,T,S) = -\log\frac{\exp(F^T_i\cdot F^S_{i}/\tau)}{\sum\limits_{j\in N} \exp(F_i^T\cdot F_j^S/\tau)} - \log\frac{\exp(F^S_i\cdot F^T_{i}/\tau)}{\sum\limits_{j\in N} \exp(F_i^S\cdot F_j^T/\tau)} ,
\end{equation}
\noindent
where $\tau$ is the temperature coefficient, similar to CLIP~\cite{radford2021learning}. Within a training mini-batch, the text-submap alignment objective $L(T,S)$ can be described as:
\begin{equation}\small
       L(T,S) = \frac{1}{N}\left [ \sum_{i\in N} l(i, T, S) \right ].
\end{equation}

{ \bf Fine localization.} Unlike previous method~\cite{kolmet2022text2pos, wang2023text}, our fine localization network does not include a text-instance matching module, making our training more straightforward and faster. Note that this model is trained separately from the global place recognition. 
Here, our goal is to minimize the distance between the predicted location of the target and the ground truth. In this paper, we use only the mean squared error loss  $L_{r}$ to train the translation regressor.
\begin{equation} 
\begin{aligned}
L(C_{gt}, {C}_{pred}) = \big \|C_{gt} - {C}_{pred}   \big \|_{2},
\end{aligned}
\label{Eq: regression loss}
\end{equation}
\noindent 
where $C_{pred}=(x, y)$ (see Eq. (\ref{Eq:encoder})) is the predicted target coordinates, and ${C}_{gt} $ is the ground-truth coordinates. 

\section{Experiments}

\subsection{Benchmark Dataset}
We train and evaluate the proposed Text2Loc on the KITTI360Pose benchmark presented in \cite{kolmet2022text2pos}. It includes point clouds of 9 districts, covering 43,381 position-query pairs with a total area of 15.51 $km^2$. Following~\cite{kolmet2022text2pos}, we choose five scenes (11.59 $km^2$) for training, one for validation, and the remaining three (2.14 $km^2$) for testing. The 3D submap is a cube that is 30m long with a stride of 10m. This creates a database with 11,259/1,434/4,308 submaps for training/validation/testing scenes and a total of 17,001 submaps for the entire dataset. For more details, please refer to the supplementary material in~\cite{kolmet2022text2pos}.

\begin{table*}[t]
    \centering
    \small
    \begin{tblr}{
      cells = {c},
      cell{1}{2} = {c=6}{},
      cell{2}{2} = {c=3}{},
      cell{2}{5} = {c=3}{},
      hline{1,7} = {-}{0.08em},
      hline{2-3} = {2-7}{0.03em},
      hline{4,6} = {-}{0.05em},
    }
                    & Localization Recall ($\epsilon < 5/10/15m $) $\uparrow$ &                                           &                                           &                                           &                                           &                                           \\
    Methods         & Validation Set                                          &                                           &                                           & Test Set                                  &                                           &                                           \\
                    & $k=1$                                                   & $k=5$                                     & $k=10$                                    & $k=1$                                     & $k=5$                                     & $k=10$                                    \\
    Text2Pos~\cite{kolmet2022text2pos}       & 0.14/0.25/0.31                                          & 0.36/0.55/0.61                            & 0.48/0.68/0.74                            & 0.13/0.21/0.25                            & 0.33/0.48/0.52                            & 0.43/0.61/0.65                            \\
    RET~\cite{wang2023text}           & 0.19/0.30/0.37                                          & 0.44/0.62/0.67                            & 0.52/0.72/0.78                            & 0.16/0.25/0.29                            & 0.35/0.51/0.56                            & 0.46/0.65/0.71                            \\
    Text2Loc (Ours) & \textbf{0.37}/\textbf{0.57}/\textbf{0.63}               & \textbf{0.68}/\textbf{0.85}/\textbf{0.87} & \textbf{0.77}/\textbf{0.91}/\textbf{0.93} & \textbf{0.33}/\textbf{0.48}/\textbf{0.52} & \textbf{0.61}/\textbf{0.75}/\textbf{0.78} & \textbf{0.71}/\textbf{0.84}/\textbf{0.86} 
    \end{tblr}
    \caption{Performance comparison on the KITTI360Pose benchmark~\cite{kolmet2022text2pos}.}
    \label{tab:fine_results}
    \vspace{-1em}
\end{table*}

\subsection{Evaluation criteria} 
Following~\cite{kolmet2022text2pos}, we use Retrieve Recall at Top $k$ ($k \in \{1, 3, 5\}$) to evaluate text-submap global place recognition.
% whereby the submap that contains the ground truth location is defined as positive. 
% In the evaluation, the best submap is labeled as positive. 
For assessing localization performance, we evaluate with respect to the top $k$ retrieved candidates ($k \in \{1, 5, 10\}$) and report localization recall. Localization recall measures the proportion of successfully localized queries if their error falls below specific error thresholds, specifically $\epsilon < 5/10/15m$ by default.

\subsection{Results}
\subsubsection{Global place recognition}
We compare our Text2Loc with the state-of-the-art methods: Text2Pos~\cite{kolmet2022text2pos} and RET~\cite{wang2023text}. 
We evaluate global place recognition performance on the KITTI360Pose validation and test set for a fair comparison. 
Table~\ref{tab:coarse_results} shows the top-1/3/5 recall of each method. The best performance on the validation set reaches the recall of 0.32 at top-1. Notably, this outperforms the recall achieved by the current state-of-the-art method RET by a wide margin of \textbf{78\%}. Furthermore, Text2Loc achieves recall rates of 0.56 and 0.67 at top-3 and top-5, respectively, representing substantial improvements of 65\% and 52\% relative to the performance of RET. These improvements are also observed in the test set, indicating the superiority of the method over baseline approaches. Note that we report only the values available in the original publication of RET. These improvements demonstrate the efficacy of our proposed Text2Loc in capturing cross-model local information and generating more discriminative global descriptors. More qualitative results are given in Section~\ref{sec: vis}.

\begin{table}[t]
    \centering
    % \small
    \resizebox{0.46\textwidth}{!}{
    \begin{tblr}{
      cells = {c},
      cell{1}{2} = {c=6}{},
      cell{2}{2} = {c=3}{},
      cell{2}{5} = {c=3}{},
      hline{1,7} = {-}{0.08em},
      hline{2-3} = {2-7}{0.03em},
      hline{4,6} = {-}{0.05em},
    }
                    & Submap Retrieval Recall $\uparrow$ &               &               &               &               &               \\
    Methods          & Validation Set                     &               &               & Test Set      &               &               \\
                    & $k=1$                              & $k=3$         & $k=5$         & $k=1$         & $k=3$         & $k=5$         \\
    Text2Pos~\cite{kolmet2022text2pos}       & 0.14                               & 0.28          & 0.37          & 0.12          & 0.25          & 0.33          \\
    RET~\cite{wang2023text}            & 0.18                               & 0.34          & 0.44          & -             & -             & -             \\
    Text2Loc (Ours) & \textbf{0.32}                      & \textbf{0.56} & \textbf{0.67} & \textbf{0.28} & \textbf{0.49} & \textbf{0.58} 
    \end{tblr}}
    \caption{Performance comparison for gloabl place recognition on the KITTI360Pose benchmark~\cite{kolmet2022text2pos}. Note that only values that are available in RET~\cite{wang2023text} are reported.}
    \label{tab:coarse_results}
    \vspace{-1.5em}
\end{table}

\subsubsection{Fine localization} 
To improve the localization accuracy of the network, \cite{kolmet2022text2pos, wang2023text} further introduce fine localization. To make the comparisons fair, we follow the same setting in ~\cite{kolmet2022text2pos, wang2023text} to train our fine localization network. As illustrated in Table~\ref{tab:fine_results}, we report the top-$k$ $(k=1/5/10)$ recall rate of different error thresholds $\epsilon < 5/10/15 m$ for comparison.  Text2Loc achieves the top-1 recall rate of 0.37 on the validation set and 0.33 on the test set under error bound $\epsilon < 5m$, which are \textbf{95$\%$} and \textbf{2 $\times$} higher than the previous state-of-the-art RET, respectively. Furthermore, our Text2Loc performs consistently better when relaxing the localization error constraints or increasing $k$.
This demonstrates that Text2Loc can accurately interpret the text descriptions and semantically understand point clouds better than the previous state-of-the-art methods. We also show some qualitative results 
in Section~\ref{sec: vis}.

\section{Performance analysis}
\label{sec: discussions}

\subsection{Ablation study}
\label{sec: ablation}
The following ablation studies evaluate the effectiveness of different
 components of Text2Loc, including both the text-submap global place recognition and fine localization. 

{\bf Global place recognition.} To assess the relative contribution of each module, we remove the frozen pre-trained large language model T5,  hierarchical transformer with max-pooling (HTM) module in the text branch, and number encoder in the 3D submap branch from our network one by one. We also analyze the performance of the proposed text-submap contrastive learning. All networks are trained on the KITTI360Pose dataset, with results shown in Table.~\ref{tab: ablation study}. Utilizing the frozen pre-trained LLM T5, we observed an approximate 8\% increase in retrieval accuracy at top 1 on the test set. While the HTM notably enhances performance on the validation set, it shows marginal improvements on the test set. Additionally, integrating the number encoder has led to a significant 6\% improvement in the recall metric at top 1 on the validation set. Notably, the performance on the validation/test set reaches 0.32/0.28 recall at top 1, exceeding the same model trained with the pairwise ranking loss by 52\% and 40\%, respectively, highlighting the superiority of the proposed contrastive learning approach.

{\bf Fine localization.} To analyze the effectiveness of each proposed module in our matching-free fine-grained localization, we separately evaluate the Cascaded Cross-Attention Transformer (CCAT) and Prototype-based Map Cloning (PMC) module, denoted as Text2Loc\_CCAT and Text2Loc\_PMC. For a fair comparison, all methods utilize the same submaps retrieved from our global place recognition.  The results are shown in Table.~\ref{tab: fine ablation study}. Text2Pos* significantly outperforms the origin results of Text2Pos~\cite{kolmet2022text2pos}, indicating the superiority of our proposed global place recognition.  Notably, replacing the matcher in Text2Pos~\cite{kolmet2022text2pos} with our CCAT results in about 7\% improvements at top 1 on the test set. 
We also observe the inferior performance of Text2Loc\_PMC to the proposed method when interpreting only the proposed PMC module into the Text2Pos~\cite{kolmet2022text2pos} fine localization network. The results are consistent with our expectations since PMC can lead to the loss of object instances in certain submaps (See Supp.). Combining both modules achieves the best performance, improving the performance by 10\% at top 1 on the test set. This demonstrates adding more training submaps by PMC is beneficial for our matching-free strategy without
any text-instance matches.

\begin{table}[t]
    \centering
    \small
    \resizebox{0.47\textwidth}{!}{
    \begin{tblr}{
      cells = {c},
      cell{1}{1} = {r=3}{},
      cell{1}{2} = {c=6}{},
      cell{2}{2} = {c=3}{},
      cell{2}{5} = {c=3}{},
      hline{1,9} = {-}{0.08em},
      hline{2-3} = {2-7}{0.03em},
      hline{4,8} = {-}{0.05em},
    }
    Methods      & Submap Retrieval Recall $\uparrow$ &               &               &               &               &               \\
                & Validation Set                     &               &               & Test Set      &               &               \\
                & $k=1$                              & $k=3$         & $k=5$         & $k=1$         & $k=3$         & $k=5$         \\
    w/o T5      & 0.29                               & 0.53          & 0.65          & 0.26          & 0.45          & 0.54          \\
    w/o HTM     & 0.30                               & 0.54          & 0.65          & \textbf{0.28} & 0.48          & 0.57          \\
    w/o CL      & 0.21                               & 0.42          & 0.53          & 0.20          & 0.36          & 0.45          \\
    w/o NE      & 0.30                               & 0.52          & 0.63          & 0.27          & 0.47          & 0.56          \\
    Full (Ours) & \textbf{0.32}                      & \textbf{0.56} & \textbf{0.67} & \textbf{0.28} & \textbf{0.49} & \textbf{0.58} 
    \end{tblr}}
    \caption{Ablation study of the global place recognition on KITTI360Pose benchmark. "w/o T5" indicates replacing the frozen pre-trained T5 model with the LSTM in ~\cite{kolmet2022text2pos}. "w/o HTM" indicates removing the proposed hierarchical transformer with max-pooling (HTM). "w/o CL" indicates replacing the proposed contrastive learning with the widely used pairwise ranking loss. "w/o NE" indicates reducing the number encoder in the instance encoder of 3D submap branch.}
    \label{tab: ablation study}
    \vspace{-1.5em}
\end{table}

% The matching-free module performs $8\%$ decrease and $8\%$ increase at the top 1 on the validation and test set, respectively. These results demonstrate that without BMC, the matching-free approach has certain limitations and may not adapt well to all scenarios. When incorporating BMC, however, the matching-free module experiences significant enhancement, reaching 0.37/0.68/0.77 on the validation set, while demonstrating slight growth on the test set.

% \begin{figure}[h]
%     \centering
%     \includegraphics[width=0.47\textwidth]{sec/vis_robust.pdf}
%     \caption{Robust analysis of our Text2Loc on the KITTI360Pose Benchmark. We present the top-3 retrieved submaps in global place recognition and the final predicted location for both the original query text descriptions and the modified queries (in red).
%     }
%     \label{fig:robust_results}
% \end{figure}

\subsection{Qualitative analysis}
\label{sec: vis}

\begin{figure*}[h]
    \centering
    \includegraphics[width=1.0\textwidth]{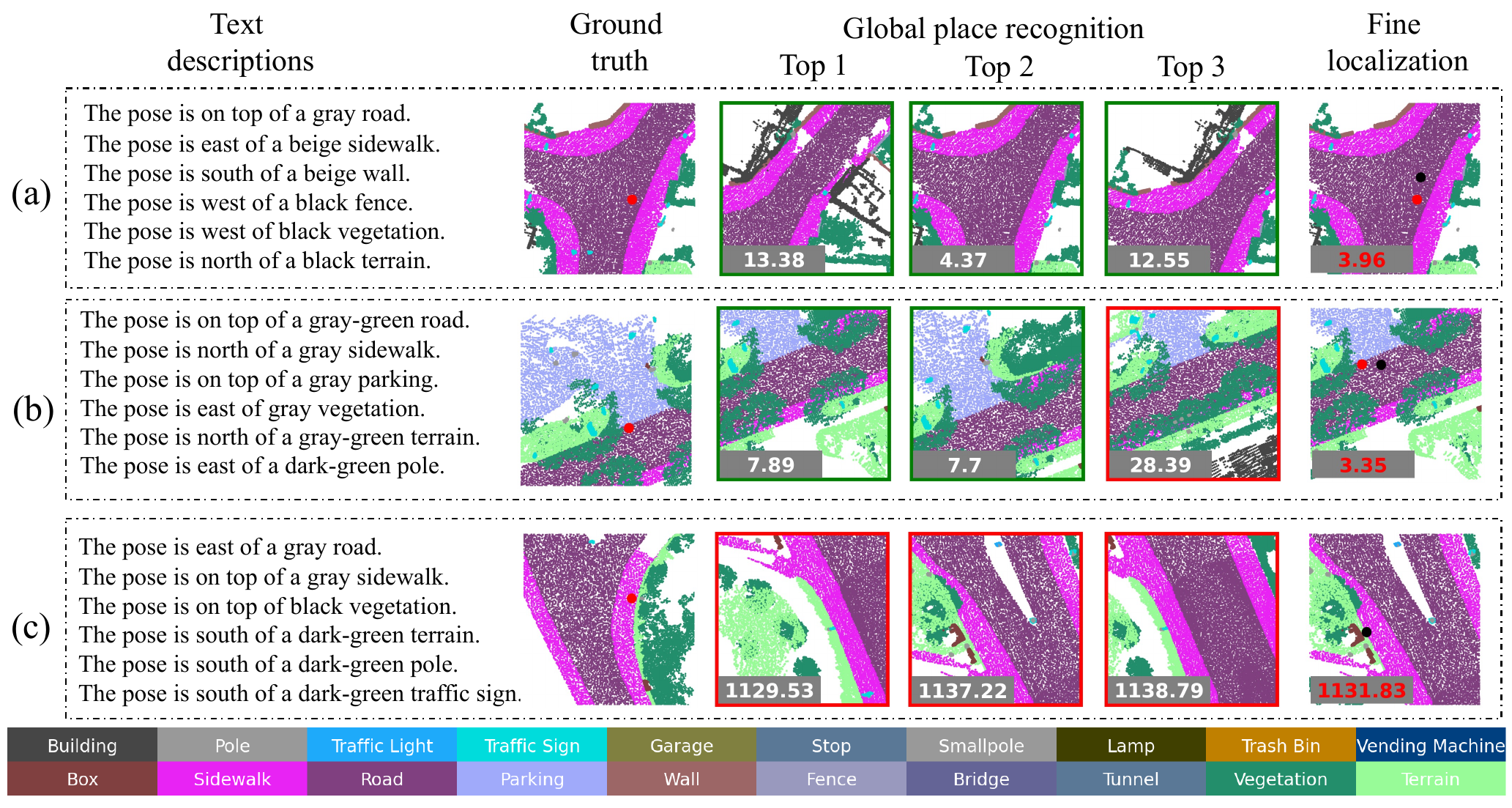}
    \caption{Qualitative localization results on the KITTI360Pose dataset: In global place recognition, the numbers in top3 retrieval submaps represent center distances between retrieved submaps and the ground truth. Green boxes indicate positive submaps containing the target location, while red boxes signify negative submaps. For fine localization, red and black dots represent the ground truth and predicted target locations, with the red number indicating the distance between them.
    }
    \label{fig:qualit_results}
\end{figure*}

In addition to quantitative results, we show some qualitative results of two correctly point cloud localization from text descriptions and one failure case in Fig.~\ref{fig:qualit_results}. 
Given a query text description, we visualize the ground truth, top-3 retrieved submaps, and our fine localization results. In text-submap global place recognition,
a retrieved submap is defined as positive if it contains the target location. Text2Loc excels in retrieving the ground truth submap or those near in most cases. However, there are instances where negative submaps are retrieved, as observed in (b) with the top 3. 
% Interestingly, even though these negative submaps are distant from the ground truth, they exhibit high semantic similarity.
Text2Loc showcases its ability to predict more accurate locations based on positively retrieved submaps in fine localization.
We also present one failure case in (c), where all retrieved submaps are negative. In these scenarios, our fine localization struggles to predict accurate locations, highlighting its reliance on the coarse localization stage.
An additional observation is that despite their distance from the target location, all these negative submaps contain instances similar to the ground truth. These observations show the challenge posed by the low diversity of outdoor scenes, emphasizing the need for highly discriminative representations to effectively disambiguate between submaps.

\begin{table}[t]
    \centering
    \small
    \resizebox{0.47\textwidth}{!}{%
    \begin{tblr}{
      cells = {c},
      cell{1}{2} = {c=6}{},
      cell{2}{2} = {c=3}{},
      cell{2}{5} = {c=3}{},
      hline{1,8} = {-}{0.08em},
      hline{2-3} = {2-7}{0.03em},
      hline{4} = {-}{0.05em},
    }
                    & Localization Recall ($\epsilon < 5m$) $\uparrow$ &               &               &               &               &               \\
    Methods         & Validation Set                                   &               &               & Test Set      &               &               \\
                    & $k=1$                                            & $k=5$         & $k=10$        & $k=1$         & $k=5$         & $k=10$        \\
    Text2Pos~\cite{kolmet2022text2pos}       & 0.14                                             & 0.36          & 0.48          & 0.13          & 0.33          & 0.43          \\                    
    Text2Pos*       & 0.33                                             & 0.65          & 0.75          & 0.30          & 0.58          & 0.67          \\
    Text2Loc\_CCAT   & 0.32                                             & 0.64          & 0.74          & 0.32          & 0.60          & 0.70          \\
    Text2Loc\_PMC    & 0.32                                             & 0.64          & 0.74          & 0.29          & 0.56          & 0.66          \\
    Text2Loc (Ours) & \textbf{0.37}                                    & \textbf{0.68} & \textbf{0.77} & \textbf{0.33} & \textbf{0.61} & \textbf{0.71} 
    \end{tblr}}
    \caption{Ablation study of the fine localization on the KITTI360Pose benchmark. * indicates the fine localization network from Text2Pose~\cite{kolmet2022text2pos}, and the submaps retrieved through our global place recognition. Text2Loc\_CCAT indicates the removal of only the PMC while retaining the CCAT in our network. Conversely, Text2Loc\_PMC keeps the PMC but replaces the CCAT with the text-instance matcher in Text2Pos.}
    \label{tab: fine ablation study}
    % \vspace{-1.5em}
\end{table}

\subsection{Computational cost analysis}
\label{sec: eff}
In this section, we analyze the required computational resources of our coarse and matching-free fine localization network regarding the number of parameters and time efficiency. 
For a fair comparison, all methods are tested on the KITTI360Pose test set with a single NVIDIA TITAN X (12G) GPU. 
Text2Loc takes \SI{22.75}{ms} and \SI{12.37}{ms} to obtain a global descriptor for a textual query and a submap respectively, while Text2Pos~\cite{kolmet2022text2pos} achieves it in \SI{2.31}{ms} and \SI{11.87}{ms}. Text2Loc has more running time for the text query due to the extra frozen T5 (\SI{21.18}{ms}) and HTM module (\SI{1.57}{ms}). Our text and 3D networks have \SI{13.65}{M} (without T5) and \SI{1.84}{M} parameters respectively. 
For fine localization, we replace the proposed matching-free CCAT module with the text-instance matcher in \cite{kolmet2022text2pos, wang2023text}, denoted as Text2Loc\_Matcher.
From Table.~\ref{tab: efficiency analysis}, we observe that Text2Loc is nearly two times more parameter-efficient than the baselines~\cite{kolmet2022text2pos, wang2023text} and only uses their 5$\%$ inference time. The main reason is that the previous methods adopt Superglue~\cite{sarlin20superglue} as a matcher, which resulted in a heavy and time-consuming process. 
% In cases with similar parameter sizes, running a transformer structure consumes less time compared to an attentional graph neural network~\cite{velikovi2017graph}. 
Besides, our matching-free architecture prevents us from running the Sinkhorn algorithm~\cite{NIPS2013_af21d0c9}. These improvements significantly enhance the network's efficiency without compromising its performance.

\begin{table}[t]
    \centering
    % \vspace{-1em}
    \resizebox{0.47\textwidth}{!}{
    \begin{tblr}{
      cells = {c},
      % vline{2-4} = {-}{},
      hline{1,4} = {-}{0.08em},
      hline{2} = {1-3}{0.03em},
      hline{2} = {4}{},
    }
    Methods         & Parameters (M) & Runtime (ms)  & Localization Recall \\
    Text2Loc\_Matcher     & 2.08           & 43.11         & 0.30                \\
    Text2Loc (Ours) & \textbf{1.06}  & \textbf{2.27} & \textbf{0.33}       
    \end{tblr}}
    \caption{Computational cost requirement analysis of our fine localization network on the KITTI360Pose test dataset. }
    \label{tab: efficiency analysis}
    \vspace{-1em}
\end{table}

\subsection{Robustness analysis}
\label{sec: robust}

In this section, we analyze the effect of text changes on localization accuracy. For a clear demonstration, we only change one sentence in the query text descriptions, denoted as Text2Loc\_modified. All networks are evaluated on the KITTI360Pose test set, with results shown in Table.~\ref{tab: robust analysis}. Text2Loc\_modified only achieves the recall of 0.15 at top-1 retrieval, indicating our text-submap place recognition network is very sensitive to the text embedding. We also observe the inferior performance of Text2Loc\_modified in the fine localization. More qualitative results are in the Supplementary Materials.

% Fig.~\ref{fig:robust_results} visualizes some qualitative results for vivid demonstration. For each instance, we display the original query text descriptions along with the top 3 retrieved submaps and their final predicted locations at the top, followed by modified queries (highlighted in red) and their results at the bottom. In the first example, we cannot find the positive submaps in the top-3 matches, leading to a complete localization failure. In the second example, even though we identify the positive submaps in the global place recognition, the exact localization is still off. The
% results are consistent with our expectation that accurate text embedding is essential for predicting the target location in fine localization. 

\begin{table}[t]
\centering
\small
\vspace{-1em}
\resizebox{0.47\textwidth}{!}{
\begin{tblr}{
  cells = {c},
  cell{1}{1} = {r=3}{},
  cell{1}{2} = {c=6}{},
  cell{2}{2} = {c=3}{},
  cell{2}{5} = {c=3}{},
  cell{3}{5} = {c=3}{},
  % vline{2} = {1-3}{},
  % vline{5} = {2}{},
  % vline{5} = {3}{},
  % vline{2,5} = {4-5}{},
  hline{1,4} = {-}{},
  hline{2-3} = {2-7}{},
  hline{6} = {1-7}{},
}
Methods             & Test set                &               &               &                     &               &               \\
                   & Submap Retrieval Recall &               &               & Localization Recall &               &               \\
                   & $k=1$                    & $k=3$           & $k=5$           & $k=5$ ($\epsilon < 5/10/15m $)                 &               &               \\
Text2Loc\_modified & 0.15                    & 0.30          & 0.38          & 0.39                & 0.54          & 0.58          \\
Text2Loc (Ours)    & \textbf{0.28}           & \textbf{0.49} & \textbf{0.58} & \textbf{0.53}       & \textbf{0.68} & \textbf{0.71} 
\end{tblr}}
\caption{Performance comparisons of changing one sentence in the queries on the KITTI360Pose test set.}
\label{tab: robust analysis}
\vspace{-1.75em}
\end{table}

\subsection{Embedding space analysis}
\label{sec: embedding space}

\begin{figure}[t]
    \centering
    \vspace{-2em}
    \includegraphics[width=0.9\columnwidth]{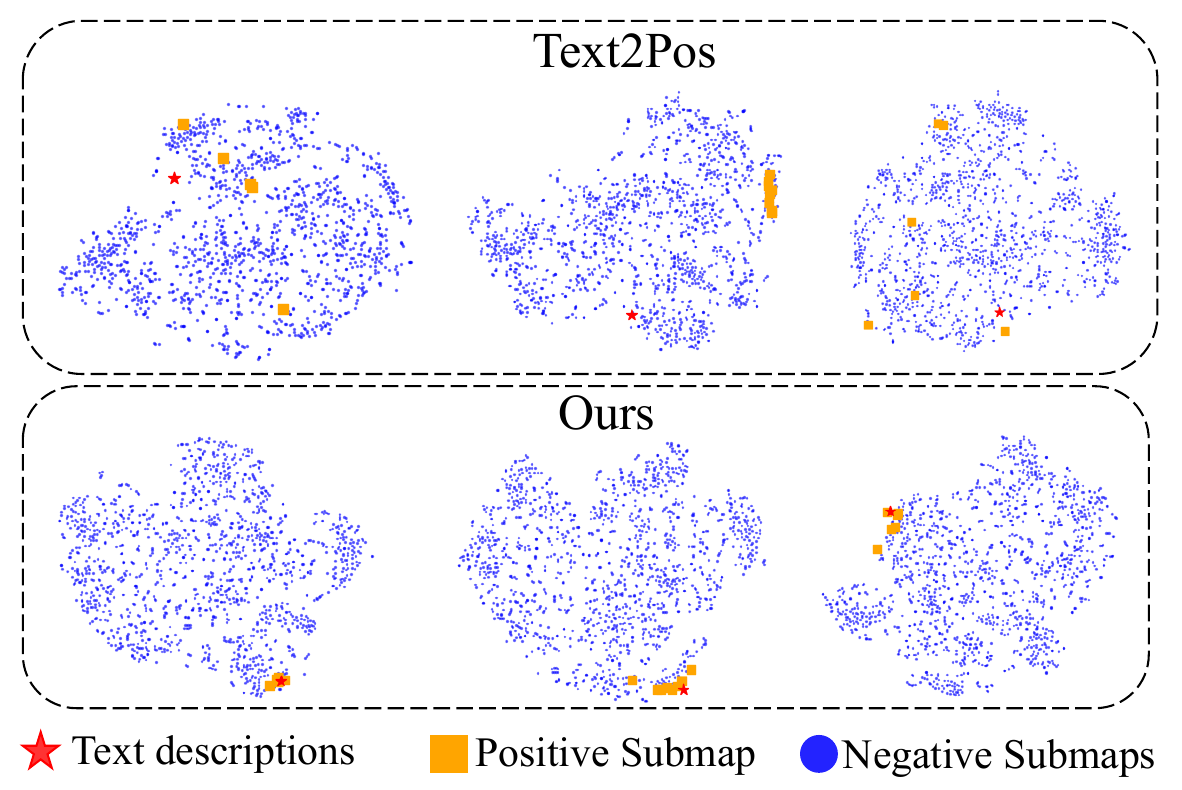}
    % \vspace{-1em}
    \caption{T-SNE visualization for the global place recognition.}
    \label{fig:tsne}
    \vspace{-3em}
\end{figure}

 % A submap is labeled as 'positive' if it contains the location mentioned in the query. Comparing our method with the baseline~\cite{kolmet2022text2pos}, our Text2Loc generates more effective representations, with positive submaps positioned closer to the target location.
We employ T-SNE~\cite{van2008visualizing} to visually represent the learned embedding space, as illustrated in Figure~\ref{fig:tsne}. The baseline method Text2Pos~\cite{kolmet2022text2pos} yields a less discriminative space, with positive submaps often distant from the query text descriptions and even scattered across the embedding space. In contrast, our method brings positive submaps and query text representations significantly closer together within the embedding distance. It shows that the proposed network indeed results in a more discriminative cross-model space for recognizing places.

\section{Conclusion}
We proposed Text2Loc for 3D point cloud localization based on a few natural language descriptions. In global place recognition, we capture the contextual details within and across text sentences with a novel attention-based method and introduce contrastive learning for the text-submap retrieval task. In addition, we are the first to propose a matching-free fine localization network for this task, which is lighter, faster, and more accurate. Extensive experiments demonstrate that Text2Loc improves the localization performance over the state-of-the-art by a large margin. Future work will explore trajectory planning in real robots.

% \vspace{0.2cm}
\noindent
{\bf Acknowledgements.} This work was supported by the ERC Advanced Grant SIMULACRON, by the Munich Center for Machine Learning, and by the Royal Academy of Engineering (RF\textbackslash 201819\textbackslash 18\textbackslash 163).

{
    \small
    \bibliographystyle{ieeenat_fullname}
    \bibliography{main}
}

\clearpage
\setcounter{page}{1}
\maketitlesupplementary
\setcounter{section}{0}
\renewcommand\thesection{\Alph{section}}

\section{Overview}
\label{sec: overview}
In this supplementary material, we provide more experiments on the KITTI360Pose dataset ~\cite{kolmet2022text2pos} to demonstrate the effectiveness of our Text2Loc and show more insights we gathered during the development. We first present thorough ablation experiments to study the impact of the proposed CCAT on the fine localization performance in Sec.~\ref{sec: as_ccat}. In Sec.~\ref{sec: vis_robust}, we provide qualitative results of top-3 candidate submaps retrieved and localization performance when changing one sequence in the query textural descriptions. Next, we describe implementation details about our network architecture in Sec.~\ref{sec: implementation} and analysis of the proposed PMC module in Sec.~\ref{sec: pmc}. Finally, Sec.~\ref{sec: vis_more} shows more visualizations of point cloud localization from text descriptions.
% 
% Having the supplementary compiled together with the main paper means that:
% % 
% \begin{itemize}
% \item The supplementary can back-reference sections of the main paper, for example, we can refer to \cref{sec:intro};
% \item The main paper can forward reference sub-sections within the supplementary explicitly (e.g. referring to a particular experiment); 
% \item When submitted to arXiv, the supplementary will already included at the end of the paper.
% \end{itemize}
% % 
% To split the supplementary pages from the main paper, you can use \href{https://support.apple.com/en-ca/guide/preview/prvw11793/mac#:~:text=Delete%20a%20page%20from%20a,or%20choose%20Edit%20%3E%20Delete).}{Preview (on macOS)}, \href{https://www.adobe.com/acrobat/how-to/delete-pages-from-pdf.html#:~:text=Choose%20%E2%80%9CTools%E2%80%9D%20%3E%20%E2%80%9COrganize,or%20pages%20from%20the%20file.}{Adobe Acrobat} (on all OSs), as well as \href{https://superuser.com/questions/517986/is-it-possible-to-delete-some-pages-of-a-pdf-document}{command line tools}.

\section{More analysis of Cascaded Cross-Attention Transformers}
\label{sec: as_ccat}
In this section, we first explore the performance of different numbers of Cascaded Cross-Attention Transformers (CCAT) in our fine localization network. We further provide a comparison to study the difference between our CCAT and Hierarchical Cross-Attention Transformer (HCAT) in ~\cite{Xia_2023_ICCV}.

\paragraph{Number of CCAT.} We insert CCAT one by one before the MLP layer in Text2Loc. '0' means using a single Cross Attention Transformer (CAT) to fuse text and 3D point cloud features. 
Table~\ref{tab: numbers of CCAT} shows the localization performance of our Tex2Loc with different numbers of CCAT units. 
As seen from the table, Text2Loc achieves the best performance with 2 CCAT units. When the number expands to 3, the performance degrades. This implies that the text-submap feature fusion is sufficient with fewer CCAT units. On the other hand, when the number is set to 1, the performance decreases. Therefore, we set the fixed number of CCAT as 2 in our network.

\paragraph{Difference with HCAT.} Recent work CASSPR~\cite{Xia_2023_ICCV} has explored the integration of 3D point-wise features with voxelized representations through a designed Hierarchical Cross-Attention Transformer (HCAT). In HCAT, two parallel Cross Attention Transformers (CAT1 and CAT2) process inputs from different branches (point and voxel), each serving as query and key respectively. In contrast, our Cascaded Cross-Attention Transformer (CCAT) employs a sequential, cascaded structure to merge text and point cloud cross-modal information. Notably, in our CCAT, the second CAT utilizes the output of the first CAT as its key and value, distinguishing it from the parallel architecture of HCAT. Table.~\ref{tab: hcat} presents a performance comparison of different modules within our Text2Loc architecture. Utilizing the proposed CCAT, we observed an 
approximate 4\% increase in retrieval accuracy at top 10 on the test set. This table demonstrates a consistently superior performance of our CCAT compared to the HCAT used in ~\cite{Xia_2023_ICCV}.

\paragraph{Motivation of CCAT.} The motivation for the CCAT module in fine localization arose from the challenge of target position regression based on the text descriptions. Encoding accurate textual features is crucial for regression since the model directly predicts target positions, without any text-instance matcher. We thus design a cascade structure to enhance text features with the information from retrieved point clouds. The HCAT~\cite{Xia_2023_ICCV} module, in contrast, aims to compensate for the quantization losses for the LiDAR-based place recognition task. HCAT should ensure that each branch is useful in isolation, thus preventing one branch from dominating over the other.

\begin{table}[]
    \centering
    \small
    \resizebox{0.47\textwidth}{!}{%
    \begin{tblr}{
      cells = {c},
      cell{1}{2} = {c=6}{},
      cell{2}{2} = {c=3}{},
      cell{2}{5} = {c=3}{},
      hline{1,8} = {-}{0.08em},
      hline{2-3} = {2-7}{0.03em},
      hline{4} = {-}{0.05em},
    }
                    & Localization Recall ($\epsilon < 5m$) $\uparrow$ &               &               &               &               &               \\
    Number of CCAT        & Validation Set                                   &               &               & Test Set      &               &               \\
                    & $k=1$                                            & $k=5$         & $k=10$        & $k=1$         & $k=5$         & $k=10$        \\
    0       & 0.28                                             & 0.57          & 0.66          & 0.26          & 0.51          & 0.60          \\
    1   & 0.36                                             & 0.67          & 0.77          & 0.32          & 0.59          & 0.69          \\
    2  & \textbf{0.37}     & \textbf{0.68} & \textbf{0.77} & \textbf{0.33} & \textbf{0.61} & \textbf{0.71} \\
    3    & 0.35      & 0.67          & 0.77     &     0.32  &        0.59          & 0.69     
    \end{tblr}}
    \caption{Localization performance for Text2Loc with different numbers of CCAT on the KITTI360Pose benchmark. '0' means using a single Cross Attention Transformer (CAT) to fuse text and 3D point cloud features. }
    \label{tab: numbers of CCAT}
    % \vspace{-1.5em}
\end{table}

\begin{table}[]
    \centering
    \small
    \resizebox{0.47\textwidth}{!}{%
    \begin{tblr}{
      cells = {c},
      cell{1}{2} = {c=6}{},
      cell{2}{2} = {c=3}{},
      cell{2}{5} = {c=3}{},
      hline{1,4,6} = {-}{},
      hline{2-3} = {2-7}{},
    }
                & Localization Recall ($\epsilon < 5m$) $\uparrow$ &               &               &               &               &               \\
    Methods     & Validation Set                                   &               &               & Test Set      &               &               \\
                & $k=1$                                            & $k=5$         & $k=10$        & $k=1$         & $k=5$         & $k=10$        \\
    HCAT~\cite{Xia_2023_ICCV}        & 0.35                                             & 0.66          & 0.75          & 0.32          & 0.59          & 0.68          \\
    CCAT (Ours) & \textbf{0.37}                                    & \textbf{0.68} & \textbf{0.77} & \textbf{0.33} & \textbf{0.61} & \textbf{0.71} 
    \end{tblr}}
    \caption{Performance comparison of different modules within our Text2Loc architecture on the KITTI360Pose benchmark. }
    \label{tab: hcat}
    % \vspace{-1.5em}
\end{table}

\begin{figure*}[htb]
    \centering
    \includegraphics[width=0.99\textwidth]{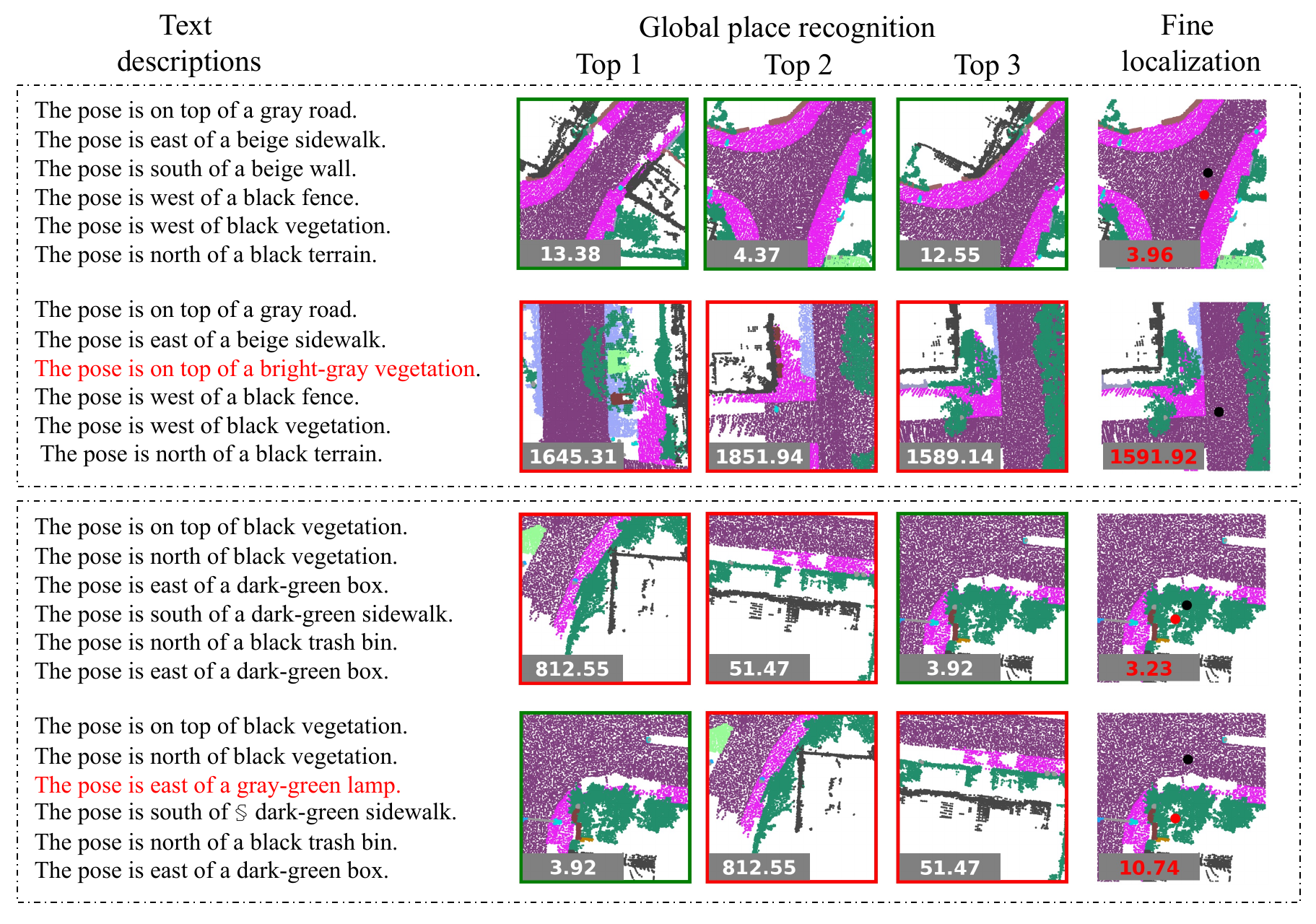}
    \caption{Robust analysis of our Text2Loc on the KITTI360Pose Benchmark. We present the top-3 retrieved submaps in global place recognition and the final predicted location for both the original query text descriptions and the modified queries (in red).
    }
    \label{fig:robust_results}
\end{figure*}

\section{Visualization of robustness analysis}
\label{sec: vis_robust}

Fig.~\ref{fig:robust_results} visualizes some qualitative results for Sec.~\ref{sec: robust}. For each instance, we display the original query text descriptions along with the top 3 retrieved submaps and their final predicted locations at the top, followed by modified queries (highlighted in red) and their results at the bottom. In the first example, we cannot find the positive submaps in the top-3 matches, leading to a complete localization failure. In the second example, even though we identify the positive submaps in the global place recognition, the exact localization is still off. The results are consistent with our expectation that accurate text embedding is essential for predicting the target location in fine localization.

% \newpage

\section{Implementation Details}
\label{sec: implementation}
We train the model with Adam optimizer for the text-submap global place recognition with a learning rate (LR) of 5e-4. The model is trained for a total 20 epochs with batch size 64, and we follow a multi-step training schedule wherein we decay LR by a factor of 0.4 at each 7 epoches. The temperature coefficient $\tau$ is set to 0.1. We consider each submap to contain a constant 28 object instances. The intra- and inter-text encoder in the text branch has 1 encoder layer respectively. We utilize PointNet++ \cite{qi2017pointnet++} from \cite{kolmet2022text2pos} to encode every individual instance within the submap. In all quantitative results relating to global place recognition, we adopt the definition of the ground truth (GT) submap as \cite{kolmet2022text2pos}, where it refers to the submap in the database that contains textual descriptions of targets, with its center point closest to the target. For the fine localization network, we train the model with an LR of 3e-4 for 35 epochs with batch size 32. To make a fair comparison, we set the embedding dimension for both text and submap branch as 256 in global place recognition and 128 in fine localization. The code is available for reproducibility. 

\paragraph{Transformer in global place recognition.} Formally, each transformer with max-pooling in the proposed intra- and inter-text encoder can be formulated as follows:
\begin{equation}
\begin{aligned}
\mathbf{F}_{T} & = \operatorname{Max-pooling \circ Transformer}(\mathbf{Q}, \mathbf{K}, \mathbf{V}) \\
& =  \operatorname{Max-pooling \circ} \left [ \widetilde{\mathbf{F}}_{T}+\operatorname{FFN}\left(\widetilde{\mathbf{F}}_{T}\right) \right ], \\
\widetilde{\mathbf{F}}_{T}& =\mathbf{Q}+\operatorname{MHSA}\left(\mathbf{Q}, \mathbf{K}, \mathbf{V}\right),
\end{aligned}
\end{equation}
\noindent
where $\mathbf{Q} = \mathbf{K} = \mathbf{V} = F_{t} \in \mathbb{R}^{N_{t} \times d}$ represent the query, key, and value matrices.

Within the MHSA layer, self-attention is conducted by projecting $\mathbf{Q}$, $\mathbf{K}$, and $\mathbf{V}$ using $h$ heads, with our choice being $h=4$.
More precisely, we initially calculate the weight matrix using scaled dot-product attention~\cite{vaswani2017attention}, as in Eq.~\ref{eq:attention}:
\begin{equation}
\operatorname{Attention}\left(\mathbf{Q}, \mathbf{K}, \mathbf{V}\right)=\operatorname{Softmax}\left(\frac{\mathbf{Q} \cdot \mathbf{K}^{T}}{\sqrt{d_{k}}}\right) \mathbf{V},
\label{eq:attention}
\end{equation}
\noindent
% where $\operatorname{S}$ denotes the Softmax function outputting attention weights.
Subsequently, we compute the values for the $h$ heads and concatenate them together as follows:
\begin{align}
\text{Multi-Head}(\mathbf{Q,K,V})=\left[ \text{ head}{1}, \ldots, \text {head}{h}\right] \mathbf{W}^{O},\\
\text {head}{i}=\text {Attention}\left(\mathbf{Q} \mathbf{W}{i}^{Q}, \mathbf{K} \mathbf{W}{i}^{K}, \mathbf{V} \mathbf{W}{i}^{V}\right),
\end{align}
\noindent
where $\mathbf{W}_{i}^{{Q,K,V,O}}$ denote the learnable parameters.

\begin{figure}[htb]
\centering
\includegraphics[width=1\columnwidth]{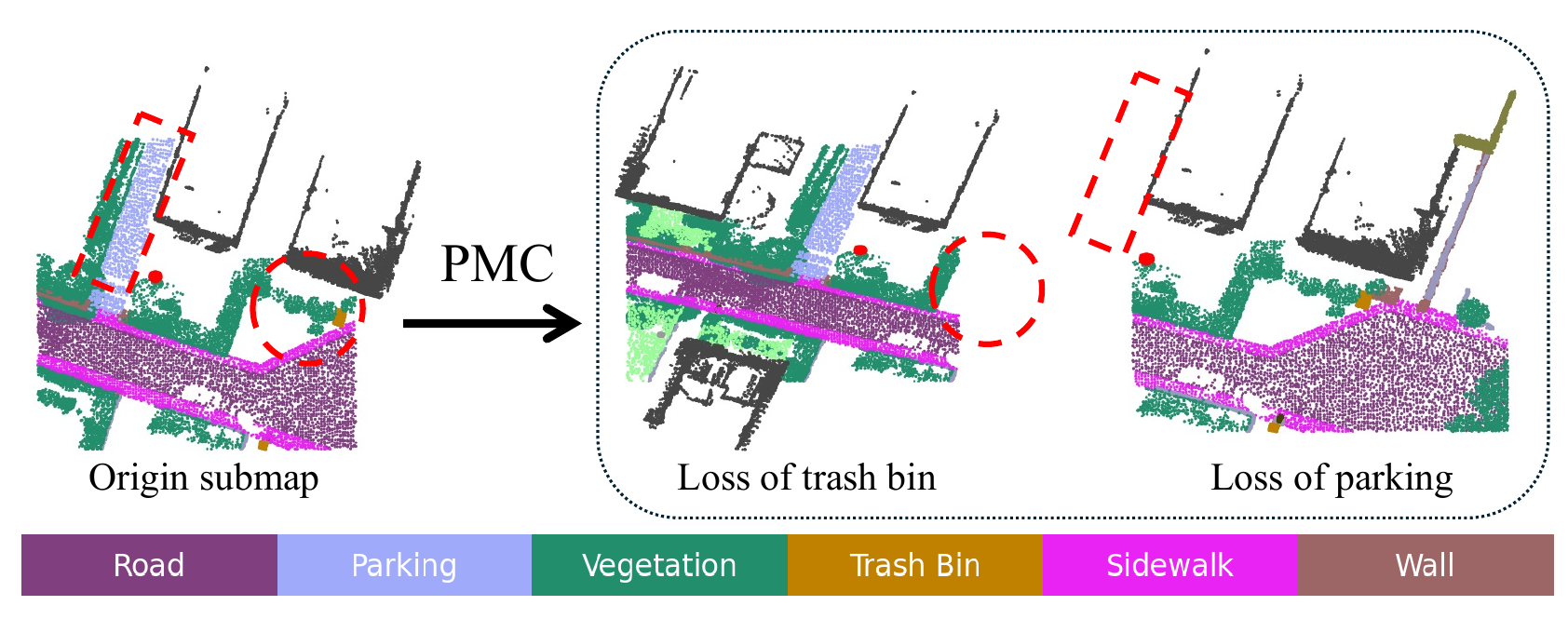}
\caption{Visualization of lost instances due to our PMC.\label{fig:voxelization}} 
\vspace{-0.4cm}
\label{fig: PMC_vis}
\end{figure}

\section{Analysis of PMC module}
\label{sec: pmc}
PMC can be seen as a data augmentation. However, this augmentation is not suitable for the previous text-instance matcher in Text2Pos~\cite{kolmet2022text2pos} and RET~\cite{wang2023text} since PMC can lead to the loss of object instances in certain submaps (see Fig.~\ref{fig: PMC_vis} above); thereby, solely integrating the PMC into Text2Pos results in performance degradation. Conversely, adding more training submaps by PMC benefits our Text2Loc since we adopt a matching-free strategy without any text-instance matches.

\section{More visualization results}
\label{sec: vis_more}
\begin{figure*}[h]
    \centering
    \includegraphics[width=0.92\textwidth]{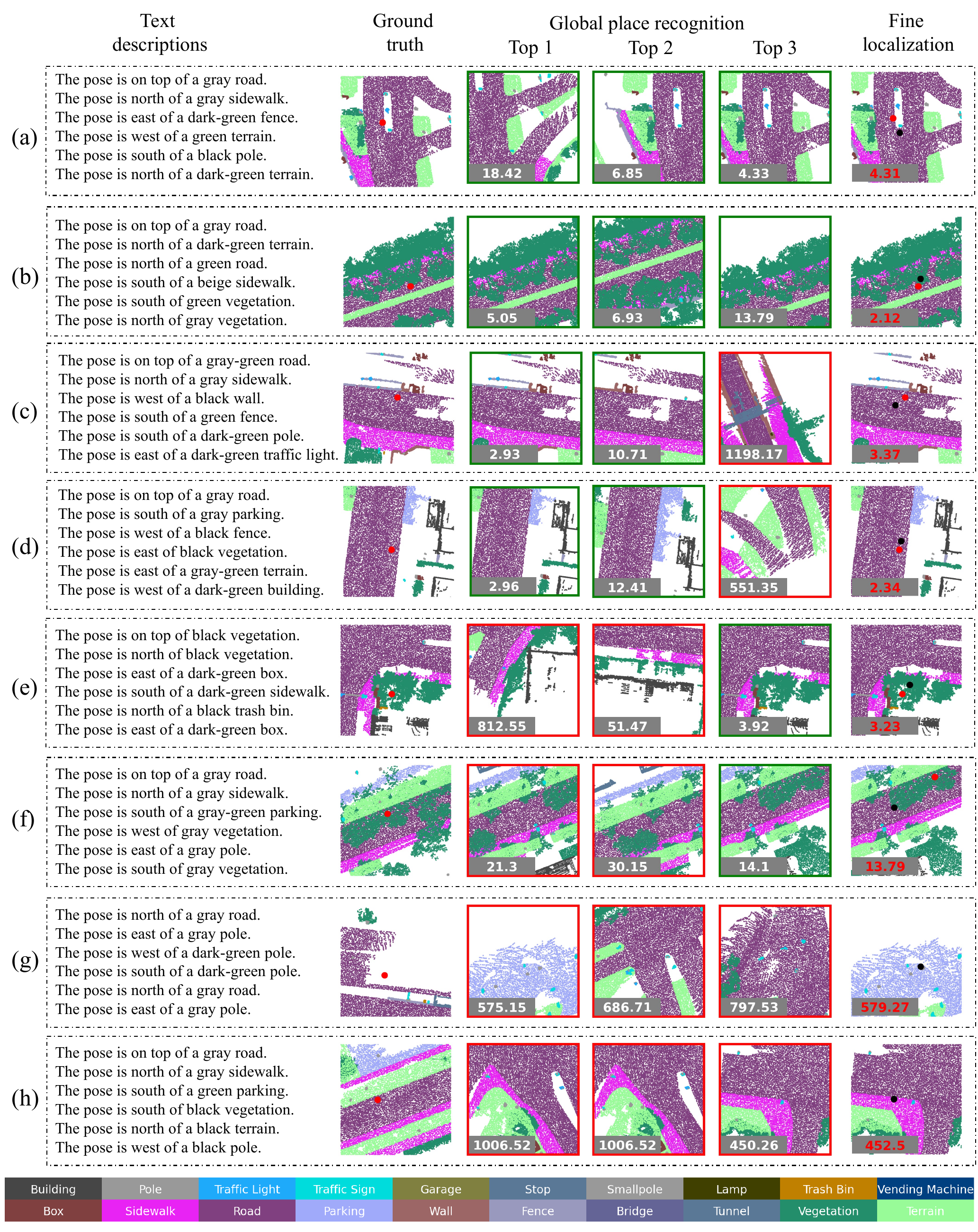}
    \caption{Qualitative localization results on the KITTI360Pose dataset: In global place recognition, the numbers in top3 retrieval submaps represent center distances between retrieved submaps and the ground truth. Green boxes indicate positive submaps containing the target location, while red boxes signify negative submaps. For fine localization, red and black dots represent the ground truth and predicted target locations, with the red number indicating the distance between them.
    }
    \label{fig:supp_qualit_results}
\end{figure*}

In this section, we visualize more examples of correct point cloud localization from text descriptions and failure cases in Fig.~\ref{fig:supp_qualit_results}. For (a) and (b), Text2Loc successfully retrieves all positive submaps within the top-3 results during global place recognition. We observe that these top-3 retrieved submaps display a high degree of semantic similarity to both the ground truth and each other. In cases of (c) - (e), despite some of the top-3 submaps being negatives retrieved by our text-submap place recognition, Text2Loc effectively localizes the text queries within a \SI{5}{m} range after applying the fine localization network. It demonstrates our fine localization network can improve the localization recall, which turns such wrong cases in place recognition into a successful localization.

We also present some failure cases where all retrieved submaps are negative. For example, in case (g), the query text description contains an excessive number of objects of the same category 'Pole'. This description ambiguity poses a significant challenge to our place recognition network, leading to the retrieval of incorrect submaps. In the future, We hope to investigate more precise and accurate text descriptions, like integrating specific landmark information, including street names, zip codes, and named buildings, into text-based localization networks.

\end{document}